\documentclass{article}

\usepackage{microtype}
\usepackage{graphicx}
\usepackage{subfigure}
\usepackage{booktabs}

\usepackage[utf8]{inputenc} 
\usepackage[T1]{fontenc}    
\usepackage{hyperref}       
\usepackage{url}            
\usepackage{booktabs}       
\usepackage{amsfonts}       
\usepackage{nicefrac}       
\usepackage{microtype}      

\usepackage{graphicx}
\usepackage{subfigure}
\usepackage{booktabs} 
\usepackage{amsmath}

\usepackage{multirow}
\usepackage{amssymb}
\usepackage{subfigure}
\usepackage{balance}
\usepackage{algorithmic}

\usepackage{hyperref}

\usepackage[accepted]{icml2019}

\icmltitlerunning{TauRieL}

\begin{document}

\twocolumn[
\icmltitle{TauRieL: Targeting Traveling Salesman Problem with a deep reinforcement learning inspired architecture}




\icmlsetsymbol{equal}{*}

\begin{icmlauthorlist}
\icmlauthor{Gorker Alp Malazgirt}{bsc}
\icmlauthor{Osman S. Unsal}{bsc}
\icmlauthor{Adrian Cristal Kestelman}{bsc}
\end{icmlauthorlist}

\icmlaffiliation{bsc}{Barcelona Supercomputing Center (BSC), Barcelona, Spain}

\icmlcorrespondingauthor{Gorker Alp Malazgirt}{gorker.alp.malazgirt@bsc.es}

\icmlkeywords{TSP, Reinforcement Learning}

\vskip 0.3in
]

\printAffiliationsAndNotice{}  

\begin{abstract}
In this paper, we propose TauRieL and target Traveling Salesman Problem (TSP) since it has broad applicability in theoretical and applied sciences. TauRieL utilizes an actor-critic inspired architecture that adopts ordinary feedforward nets to obtain a policy update vector $v$. Then, we use $v$ to improve the state transition matrix from which we generate the policy. Also, the state transition matrix allows the solver to initialize from precomputed solutions such as nearest neighbors. In an online learning setting, TauRieL unifies the training and the search where it can generate near-optimal results in seconds. The input to the neural nets in the actor-critic architecture are raw 2-D inputs, and the design idea behind this decision is to keep neural nets relatively smaller than the architectures with wide embeddings with the tradeoff of omitting any distributed representations of the embeddings. Consequently, TauRieL generates TSP solutions two orders of magnitude faster per TSP instance as compared to state-of-the-art offline techniques with a performance impact of 6.1\% in the worst case.
\end{abstract}

\section{Introduction}
\label{sec:intro}
\begin{figure}
\centering     
\subfigure[]{\label{fig:a}\includegraphics[width=0.534\linewidth]{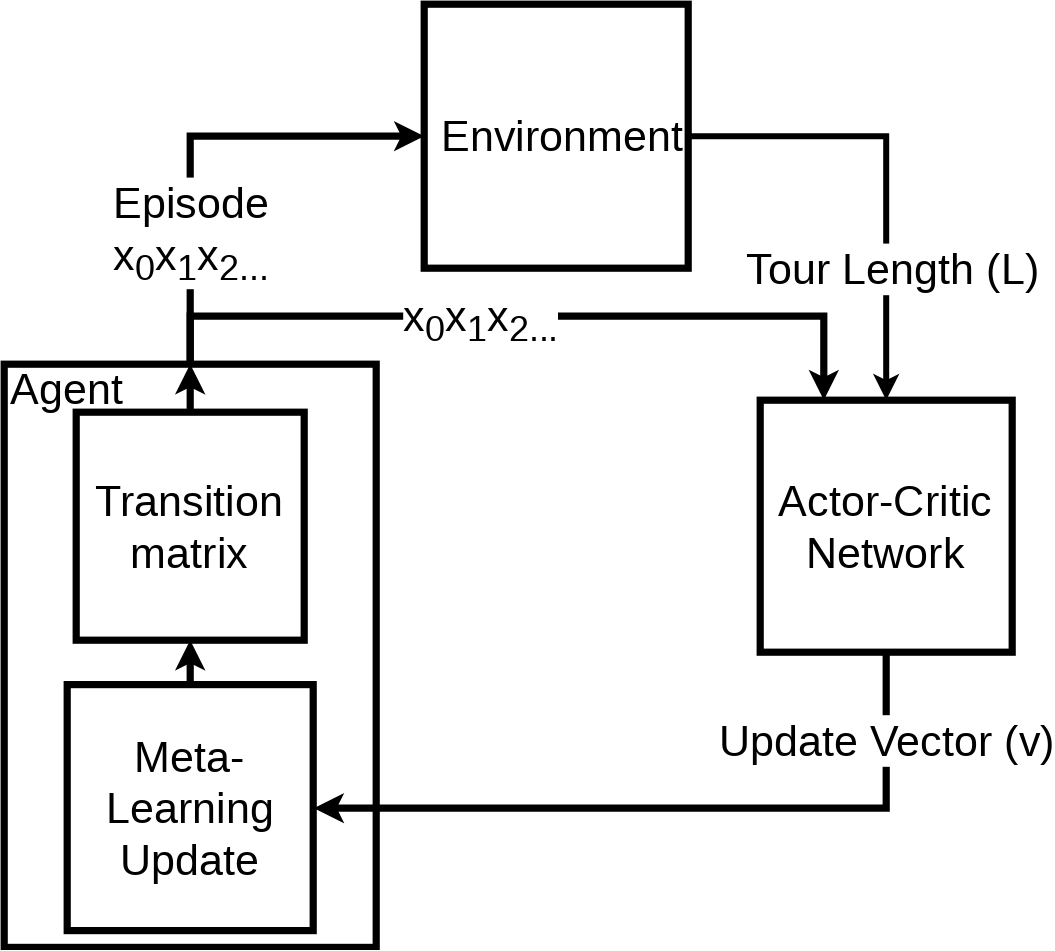} \label{fig:arch}}
\subfigure[]{\label{fig:arch_other}\includegraphics[width=0.53\linewidth]{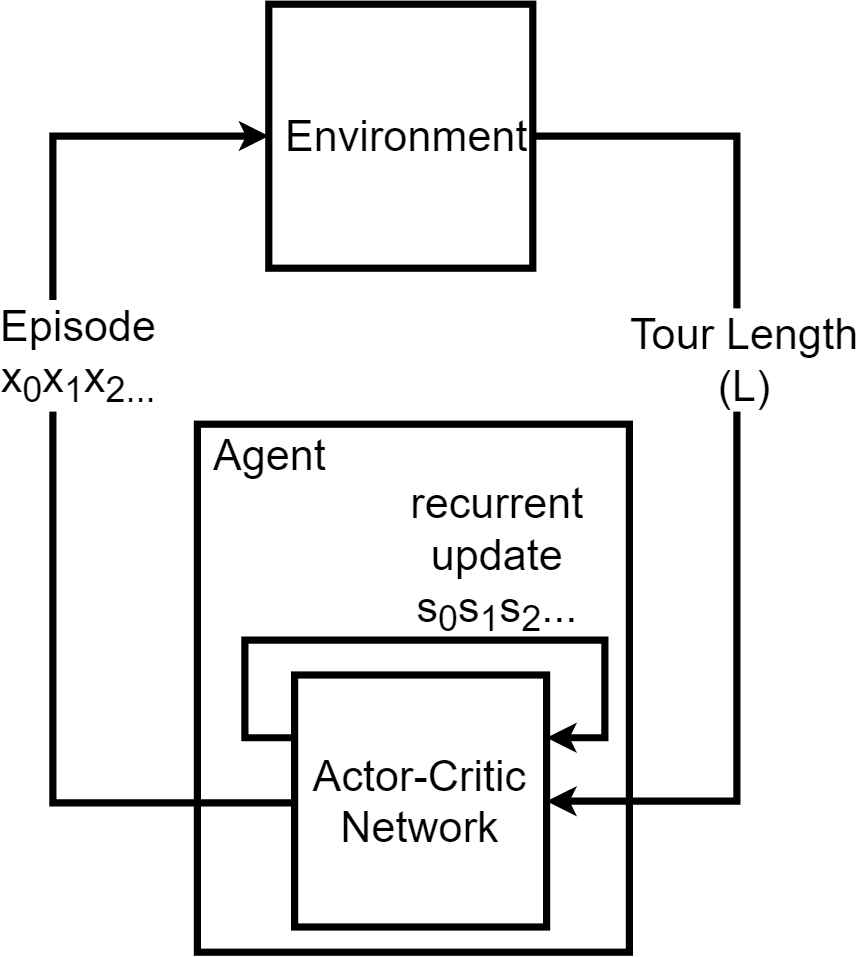} \label{fig:archelse}}
\caption{High level schemas of TauRieL (a) and a state-of-the-art Actor-Critic based TSP solver using RNN \cite{bello2016neural}(b)}
\end{figure}

In this paper, we introduce TauRieL \footnote{A wood-elf character from Hobbit the Movie who possesses \textit{superior pathfinding abilities}}; a Deep Reinforcement Learning (DRL) inspired TSP solver. Figure \ref{fig:arch} presents the flow of TauRieL. In the TauRieL architecture, there exists an agent that is responsible for making decisions by sequentially taking actions. In this setting, taking actions leads to creating traveling salesman tours. 

Traveling Salesman Problem (TSP) is one of the most well known NP-Hard problems in the fields of computer science and operations research \cite{karp1972reducibility}. TSP seeks for the shortest tour of a salesman visiting multiple cities, each exactly once. Given a graph of cities where the cities are the nodes and the edge costs represent the distance between cities, the problem can be formulated as finding the shortest tour length among all the city permutations which are the sequences of cities that the salesman visits once. Thus, the optimum sequence/tour is the permutation that creates the minimum total edge costs (tour length). 

TSP has been formulated and studied widely as a combinatorial optimization problem, and approximate heuristics based methods have been proposed. However, the design of heuristics requires detailed domain-specific knowledge \cite{stutzle2017iterated} and tuning. Compared to these heuristics, neural network (NN) based TSP solvers are more generic to develop and to adapt to different TSP variants \cite{goodfellow2016deep}. In this paper, we attack this aspect and develop a novel NN-based TSP method that can produce feasible results two orders of magnitude faster than current NN-based methods with competitive solution quality.

TauRieL employs actor-critic inspired architecture and represents the actor and the critic with feedforward nets \cite{sutton2018reinforcement}. Coupled with the actor-critic architecture, we introduce the transition matrix. The transition matrix represents the agent's view of the environment. Specifically, the agent's view of the environment is the transition probabilities between the cities. Thus, the agent acts stochastically according to the transition matrix and searches the design space. We define the final policy as the deterministic mapping that indicates which city to travel to next, given a city. Also, the state transition matrix allows TauRieL to initialize from precomputed solutions such as the nearest neighbors heuristic \cite{stutzle2017iterated}.

In the actor-critic inspired architecture, TauRieL iteratively updates its view of the environment by generating an update vector $v$ through two neural networks representing the actor and the critic \cite{sutton2018reinforcement}. Specifically, the actor-network is responsible for generating the update vector $v$ that updates the transition matrix. The critic network is responsible for estimating the Euclidean tour length from a given tour, and its primary duty is to improve the agent's exploration of the design space. We optimize the parameters of the actor and the critic model by adopting the REINFORCE and stochastic gradient descent algorithms respectively \cite{williams1992simple, robbins1951stochastic}.

TauRieL takes raw inputs, and the design idea behind this decision is to keep neural net sizes relatively small for online learning and solving TSP instances. In addition, TauRieL omits embeddings with the tradeoff of omitting any distributed representations of the embeddings. The current methods employ offline training and wide embeddings \cite{vinyals2015pointer,bello2016neural,khalil2017learning,deudon2018learning,emami2018learning}.

Deep learning based combinatorial optimization heuristics has challenges \cite{bengio2018machine}. First, unlike integer programming or constraint optimization methods \cite{baldick2006applied}, neural network based heuristics do not provide any clues regarding how close the solutions are from the optimal or do not dictate any means of feasible candidate solution generations. Thus, researchers must design architectures that can generate feasible solutions. Second, unlike vision or natural language processing problems, the characteristics of the distribution at hand for a combinatorial problem is usually unknown. Thus, the generalization from an unknown distribution could only be possible for application-specific use cases or small scale datasets.

TauRieL addresses the mentioned challenges in the following ways: First, similar to Nazari et al. \cite{nazari2018reinforcement} it enforces the generation of feasible solutions through conveying constraints in the transition matrix by nulling infeasible transitions. Second, similar to Bello et al. \cite{bello2016neural} TauRieL addresses the generalization challenge by intermixing the training and the pathfinding tasks in search of the optimal TSP tour per TSP instance.

Furthermore, our contributions in this work are listed as:
\begin{itemize}
\item We present an online training based DRL TSP solver in a deep reinforcement learning setting built for rapid exploration of the design space given a TSP instance.
\item We implement a learnable update method that interconnects actor-network and the transition matrix .
\item We compare TauRieL's findings with state-of-the-art and provide a detailed breakdown of execution times of each TauRieL building block. 
\end{itemize}

Experimental results demonstrate that TauRieL generates TSP solutions two orders of magnitude faster per TSP instance as compared to state-of-the-art offline techniques with a performance impact of 6.1\% in the worst case for 50-city instances.

We continue forward with explaining the notations and describing the TSP in the next section. The rest of the next section illustrates the building blocks of TauRieL shown in Figure \ref{fig:arch}. Section 2.2 explains TauRieL's Actor-Critic intrinsics; Section 2.3 discusses how episodes are generated and initialized in the transition matrix. Section 2.4 describes how we update the transition matrix. Section 3 introduces the pseudo-algorithm and Section 4 presents the experimental results. Section 5 discusses related works, and we finalize the paper with the conclusions and the future work.

\section{Reinforcement Learning Method for TSP}
\label{sec:RLTSP}
\subsection{Problem Definition and Notations}
\label{subsec:defs}
Given a graph $G$ that consists of cities $G=\{x\}_{i}^n$ such that $x_{i} \in \mathbb{R}^2$, the objective is to find shortest tour by visiting each city \cite{lin1973effective}. We represent the environment as a Markov Decision Process (MDP), which is a tuple $\langle S, A, P, R, \gamma \rangle$. $S$ defines a state space where each state $s$ consists of a city $x \in \mathbb{R}^2 \land x\subseteq G$. 

$P$ is a state transition probability matrix and $\mathit{P_{i,j}} = \mathit{P(S_{t+1}}=s_{j} \mid S_{t}= s_{i})$ such the probability of reaching state $s_{j}$ at time $t+1$ from $s_{i}$ at time $t$ is $P_{i,j}$. We define $R$ as the expected reward $R_{s} = \mathbb{E}[R_{t+1} \mid S_{t} = s]$ which presents the expected reward that the agent obtains when reaching state $s$ , $\gamma$ is the discount factor $ \gamma \in (0,1)$. We define the reward $r_{i,j}$ as the negative distance between two cities $x_{i}$ and $x_{j}$. 

\begin{figure}[t] 
\vskip 0.1in
\begin{center}
\centering
\centerline{\includegraphics[scale=0.18]{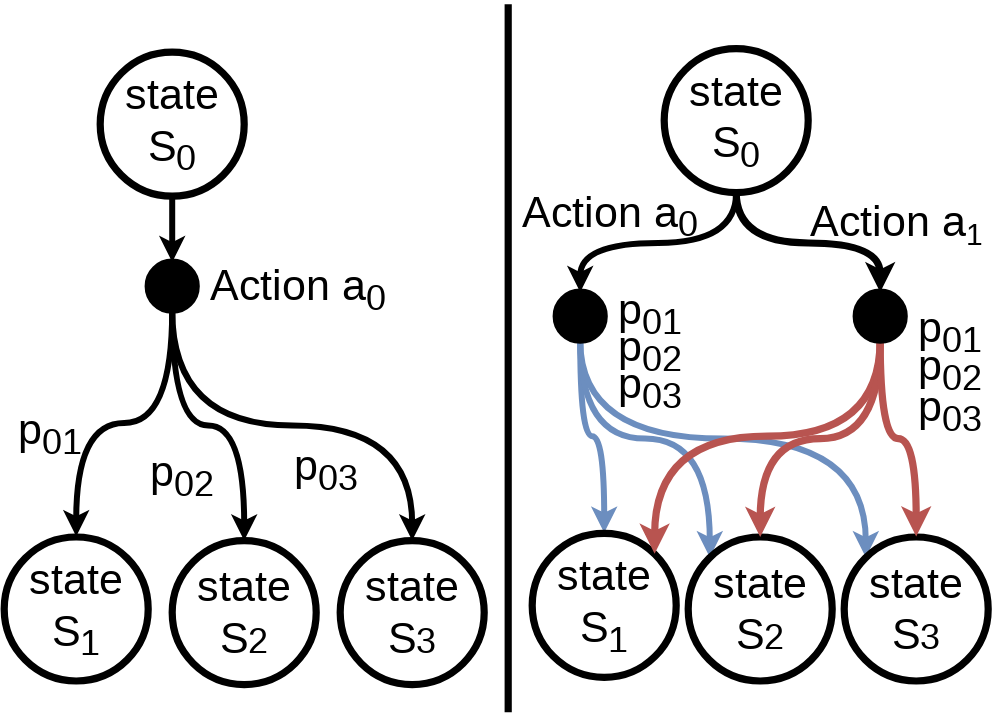}}
\caption{In TSP, there exists only a single set of action to move from one state to another (left). In a video game there are scenarios where at each state there can be more than one action such as directions of controller movements (right)}
\label{fig:action}
\end{center}
\vskip -0.2in
\end{figure}

$A$ is the set of actions $\{a_{1}, a_{2}\dots\}$. An example of a set actions can be the different directions of controller movements in a video game \cite{mnih2013playing}. For TSP, we assume that the cardinality of the action set is one and this action moves the agent between states which we visualize in Figure \ref{fig:action}. Therefore for TSP,  we assume that each state is synonymous with the action. For example, from state $s_{0}$ taking action $a_{0}$ will transition to a new state $s_{i}$ with probability distribution in $\mathit{P_{0,i}}$. We describe the policy as $\pi: S \times A \mapsto S$ and the policy $\pi(a_{0} \mid s_{t})$ transitions to a new state $s_{t+1}$ according to the dynamics $P(s_{t+1}|s_{t},a_{0})$ and receives a reward $r(s_{t+1} \mid s_{t},a_{0})$. 

Given the graph $G$ and policy $\pi$ a tour $\phi$ then represents city traversals. The environment observes the episode i.e. a valid tour and returns the total reward as the length $L$ of the tour. Thus, given a graph, we define $L$ as:

\vspace{-1cm}

\begin{equation}
L(\phi \mid G)\,=\,\Vert x_{\phi(n)}-x_{\phi(1)}\Vert_{2}+\sum_{i=1}^{n-1}\Vert x_{\phi(i)}-x_{\phi(i+1)}\Vert_{2} 
\end{equation}

We compute the probability of a tour through the chain rule: 

$p(\phi \mid G)=p(\phi_{1})p(\phi_{2}|\phi_{1})\ldots p(\phi_{n-1}\,|\,\phi_{n-2})$.

Note that tours which generate higher rewards (where L is small) will have higher $p(\phi \mid G)$.

Whenever a problem can be formulated with a chain rule such as TSP or natural language processing such that a chain of conditional probabilities expresses the probability of a string as \cite{brown1992class}, recurrent models \cite{sutskever2014sequence} and more recently sequence-to-sequence models with attention \cite{vinyals2015pointer,lecun2015deep} are apt candidates for the problem at hand. The advantage of these approaches is that the model output refers to one of the input elements rather than to a fixed set of reference inputs such as a language model \cite{DBLP:journals/corr/WuSCLNMKCGMKSJL16}.

\subsection{Using Actor-Critic Reinforcement Learning to Generate the Update Vector}
\label{sebsec:actcri}
In this section, we present the internals of the Actor-Critic building block of Figure \ref{fig:arch}. The goal is to optimize the parameters $\theta_{act}$ of the neural net that yields the best policy update vector $\textit{v}$. We optimize the parameters of the neural net with respect to the objective i.e. the expected tour length: 

\vspace{-0.7cm}

\begin{equation} 
\label{eqn_j}
J(\theta_{act}|G)=\mathbb{E}_{\phi \sim p(. \mid s)}L(\phi\mid G)
\end{equation} 

This neural net is called the actor because the gradients with respect to parameters are updated in the direction of improving the update vector $v$. The gradient $\nabla_{\theta_{act}}$ of the expected tour length is calculated using the REINFORCE algorithm \cite{williams1992simple} that we tailored for TSP as shown below:
\vspace{-0.5cm}

\begin{equation} 
 \label{eqngrorg}
\begin{aligned}
\nabla_{\theta_{act}}J(\theta_{act}|G) & =\mathbb{E}_{\phi \sim p(. \mid s)}[(L(\phi\mid G)-b(G))\\
& \quad  \; \nabla_{\theta_{act}}log\;p(\phi \mid G)]
\end{aligned}
\end{equation}

Using a stochastic batch gradient method, we estimate the gradient from batches that are sampled from the transition matrix:  

\vspace{-0.5cm}

\begin{equation} 
\label{eqngr}
\begin{aligned}
\nabla_{\theta_{act}}J(\theta_{act}|G) & =\frac{1}{B}\sum_{i=1}^{B}[(L(\phi \mid G)-b(G))\\
& \quad \; \nabla_{\theta_{act}}log\; p(\phi \mid G)]
\end{aligned}
\end{equation}

In the actor-critic architecture, the baseline $b(G)$ is a parametric metric that all generated tour lengths can be compared against. Hence, we use $b(G)$ in REINFORCE to reduce large fluctuations of the tour lengths observed by the agent during the search. In this work, we select the baseline as the estimated tour length value obtained from the critic. Therefore, we adopt a neural net for critic that approximates the expected tour length from a given path. The critic net evaluates the current policy by estimating the expected tour length and aims to prescribe toward improved tours. We train the parameters of the critic $\theta_{cri}$ using stochastic gradient descent on a mean squared error objective $H$ between its predictions and the tour length that we obtain from the most recent episode:

\vspace{-0.5cm}

\begin{equation} 
\label{eqn_length}
H(\theta_{cri}|G)=\frac{1}{B}\sum_{i=1}^{B}(b(G) - L(\phi \mid G))^{2}
\end{equation} 

Although baseline $b(G)$ is independent of the final policy $L(\phi \mid G)$, the training of the critic network and the actor network occurs concurrently. Both the actor and the critic receive raw input vectors of episodes; the actor outputs the update vector $v$ and the critic outputs a tour length estimation which is represented as the baseline. We detail the scheme that updates the transition matrix in the next section.

\subsection{Sampling from the transition matrix}
\label{subsec:sampling}
In this section, we explain how to sample episodes from the transition matrix shown in Figure \ref{fig:arch}. The sampling of episodes from the transition matrix occurs at each step. Thus the policy $\pi(a_{0}|S_{t}=s_{i})$ transitions to a new state $s_{j}$ with probability $P_{i,j}$ in the state transition matrix.
 
Each row of the state transition matrix $P_{i,:} \; i \in {i, \ldots n}$ represents a probability distribution. Thus, creating a permutation $\phi$ from $P$ can be generalized by defining a function which receives a distribution such as $P$ and returns a permutation. 

For each city, we choose the most likely city from the transition matrix $f(s) = \{argmax_{p} P_{i,:}\}$. If the most likely city has been chosen previously, the next mostly likely available city is selected. The final policy is also determined similarly.

Transition matrix also allows embedding precomputations. For example, for a given instance, a precompute step could be to use a greedy TSP heuristic such as nearest neighbor search \cite{glover2006handbook}. Thus, in this case we initialize the starting transition probabilities by leveraging the transitions selected by the precompute step. For example, in a four-city TSP case with cities $a,b,c,d$, if the greedy heuristic decides on the transitions $P_{a,c}$, $P_{c,b}$ and $P_{b,d}$ as the shortest tour, we initialize the transition matrix with higher values compared to the rest of the transitions. We show the valid episodes that we feed into the actor-critic in Figure \ref{fig:arch}.

\begin{algorithm}[tb]
 \caption{Pseudo algorithm of TauRieL}
 \label{alg:search}
\begin{algorithmic} [1]
\STATE {\bfseries Input:} input graph $G$, number of search steps $steps$, batch size $B$, episode samples $T$, learning rate $\epsilon$, update steps $K$

\STATE {\bfseries Output:} the shortest tour length  $L_{min}$, the policy that yields the shortest length $\pi$, state transition matrix $P$
\STATE Initialize actor and critic neural net parameters $\theta_{act}$ and $\theta_{cri}$
\STATE Initialize the transition matrix $P$
\STATE $\phi \leftarrow RandomEpisode(G)$
\STATE $L_{\pi} \leftarrow L(\phi \mid G)$
\FOR {t=1,...,$steps$}
\STATE $\phi_{i} \gets \:  SampleEpisodes(P(. \mid S = s_{i} ))$ for $i \in \left\{ 1,\ldots ,B\right\} $ (Sample from Transition Matrix given start state)
\STATE $j \leftarrow argmin(L(\phi_{1}),\ldots,L(\phi_{B}))$ (Shortest tour)
 \IF {$L_{j} < L_{\pi}$}
 \STATE  $L_{\pi} \leftarrow L_{j} $
 \STATE  $\pi \leftarrow \pi_{j}  $  
 \ENDIF

\STATE $\nabla_{\theta_{act}}J(\theta_{act}|G)=\frac{1}{B}\sum_{i=1}^{B}[(L(\phi \mid G)- b(G))$
\\\hspace{2.6cm} $\nabla_{\theta_{act}}log\; p(\phi \mid G)]$ (Eq. 4)
\STATE $H(\theta_{cri}|G)=\frac{1}{B}\sum_{i=1}^{B}(b(G) - (L(\phi \mid G))^{2}$ (Eq. 5)
\STATE $\theta_{act} \leftarrow RMSProp(\theta_{act},\nabla_{\theta_{act}}J(\theta_{act}|G))$
\STATE $\theta_{cri} \leftarrow RMSProp(\theta_{cri},\nabla_{\theta_{cri}}H(\theta_{cri}|G))$
\STATE $v \leftarrow p(\phi \mid G)$ 
 \IF {$K$ steps}
\STATE $P_{i,j}=P_{i,j}+ \epsilon \; (v_{i} - P_{i,j})$ (Eq. 6)
 \ENDIF
\ENDFOR
\end{algorithmic}
\end{algorithm}

\subsection{Learning to update the transition matrix}
\label{subsec:update}
The actor net in the actor-critic architecture in Figure \ref{fig:arch} is responsible for producing the update vector $v$ and after each $K$ episode the transition matrix is updated with $v$. Thus, in this section we explain the transition matrix update procedure which corresponds to the Meta-Learning Update building block in Figure \ref{fig:arch}.

We treat this step as updating the parameters of a neural net towards the final parameters learned on a task through a gradient descent optimization algorithm \cite{bengio2013advances}. We present the update in Equation \ref{eqn:delta}: 

\begin{equation}
\label{eqn:delta}
\begin{aligned}
P_{i,j} & = P_{i,j} + \epsilon \; (v_{i} - P_{i,j}) \\ 
& \quad \; \forall i \; [i \in {1,\ldots, n}] \; and \; \exists j \; [ j \in {1,\ldots, n}]
\end{aligned}
\end{equation}  

The update vector $v$ contains $n$ elements, representing each city in the permutation $\phi$. If each element of the permutation is generated from the transition matrix, then each transition $P_{i,j} \forall i \in {1,\ldots, n}$ is sampled via $f(s)$ as previously defined. The main design idea of varying the $K$ is to allow sampling from $P$ more than just one step, and it allows more exploration at the current version of the state transition matrix before an update. Additionally, this provides the algorithm to gradually increase $K$ towards the later stages for allowing early exploration. The learning parameter $\epsilon$ is a hyperparameter, and we perform a grid search to optimize it \cite{goodfellow2016deep}.

\section{Unified Training and Searching for the Shortest Tour}
The pseudo algorithm of TauRieL for finding the shortest tour is presented in Algorithm \ref{alg:search}. The algorithm presents our approach of unified training and searching. Line 3-4 initializes the actor and critic nets and the transition matrix. The transition matrix can either be initialized randomly or from a predetermined initialization that exerts explicit rules. For example, it is possible to prevent certain transitions between states by assigning corresponding probabilities to zero (Line 4). Line 5 generates a random episode and stores the tour length (Line 6).

The search step starts with sampling episodes from the transition matrix (Line 8). Then the shortest tour is obtained among the samples (Line 9). If the obtained tour is the shortest so far, it is assigned as the current minimum and the policy $\pi$ is updated based on the current minimum tour (Line 10-12). Next, the Actor gradient approximation and the Critic loss are calculated (Line 14-15), and the Actor and Critic nets are forward propagated in this process that are shown with $b(\phi \mid G)$ and $p(\phi \mid G)$. After the backward passes of the actor-critic net (Line 16-17) the transition matrix update occurs after $K$ steps (Line 20) using the update vector $v$ (Line 18). The algorithm returns the minimum tour length, the policy that generates the tour lengths and the transition matrix $P$.

\section{Experimental results}
\label{sec:exp_res}
We present our results in this section. We generate all experiments by applying our methods from the previous sections. We use Tensorflow \cite{abadi2016tensorflow} framework for all the software implementations.

All experiments run on computing nodes with the following specs: Intel Xeon E5–2630@2.4 GHz, Nvidia K80 and 64GB DDR4@2133MHz RAM. We have experimented with 20 and 50-city instances of TSP and used the dataset from \cite{vinyals2015pointer} as well as uniformly generated random points $[0,1] \in \mathbb{R}^2$.

In all the experiments, actor and critic neural nets take mini-batches of 4 instances. For the actor net, we use a 6-layer feedforward net with:
\vspace{-0.2cm}

$[64,32,32,32,32,number\_of\_cities]$ neurons. For the critic neural net, we use 5-layer feedforward net with $[64,32,32,16,8,1]$ neurons. For both nets, we use ReLu activations and RMSProp optimization algorithm \cite{tieleman2012lecture}. The configurations are the following: The learning rates are set to $3\mathrm{e}{-4}$ and $2\mathrm{e}{-4}$ for actor and critic, decay is set to $0.96$ and epsilon as $1\mathrm{e}{-6}$. The actor net outputs a vector of $v$ of size $n$ and the critic net outputs a floating point scalar estimating the tour length given a permutation.

\begin{table}[t]
\caption{Comparison of average tour lengths using the datasets provided by Ptr-Net \cite{vinyals2015pointer}, NCO \cite{bello2016neural},  Sinkhorn Policy Gradient (SPG) \cite{emami2018learning} and A3 algorithm \cite{a3algo} obtained from \cite{vinyals2015pointer}}
\label{tbl:length}
\vskip 0.15in
\begin{center}
\begin{small}
\begin{sc}
\begin{tabular}{lccccccr}
\toprule
n & Optimal & A3 & Ptr-Net & NCO & SPG& Ours \\
\midrule
10 & 2.87 & 3.07 & 2.87 & NA & NA & 2.88 \tabularnewline
20 & 3.82 & 4.24 & 3.88 & 3.96 & 4.62 & 3.91 \tabularnewline
50 & 5.68 & 6.46 & 6.09 & 5.87 & NA & 6.37 \tabularnewline
\bottomrule
\end{tabular}
\end{sc}
\end{small}
\end{center}
\vskip -0.1in
\end{table}

The Algorithm \ref{alg:search} requires four hyperparameters; these are the number of iterations $steps$, learning rate $\epsilon$, update steps $K$ for state transition matrix update and sample steps $T$. In all the presented results, the number of iterations $steps$ and the sampling $T$ are set to 250. The learning rate $\epsilon$ is set to 0.01.

\begin{table}[t]
\caption{Execution times in seconds of a single episode and sample step for TSP20 and TSP50 instances}
\label{tbl:execution_time}
\vskip 0.15in
\begin{center}
\begin{small}
\begin{sc}
\begin{tabular}{lll}
\toprule 
 & N = 20 & N = 50\tabularnewline
\midrule
Graph Gen. & 0.014 & 0.023\tabularnewline
Training & 0.0034 & 0.0035\tabularnewline
Sampling & 0.00123 & 0.0018\tabularnewline
Inference & 0.00102 & 0.0012\tabularnewline
Transition Matrix Up. & 0.0003 & 0.0004\tabularnewline
Cl Merge & 0.007 & 0.006 \tabularnewline
Other & 0.0005 & 0.0005\tabularnewline
\bottomrule
\end{tabular}
\end{sc}
\end{small}
\end{center}
\vskip -0.1in
\end{table}

We compare our results with the tour lengths obtained from A3, Ptr-Net, NCO, and Sinkhorn Policy Gradient \cite{a3algo,vinyals2015pointer, bello2016neural,emami2018learning}. Table \ref{tbl:length} presents average tour lengths for 10 to 50-city instances. TauRieL is within 1\% of the optimal tour for the 10-city. TauRieL outperforms A3 for both 20 and 50 city instances and obtains tour lengths within 0.007 \% and 2\% of Ptr-Net for 20 and 50-city instances. Apart from 50-city instances, Ptr-Net is trained with optimal results \cite{vinyals2015pointer}. 

Moreover, TauRieL outperforms an actor-critic based Sinkhorn Policy Gradient \cite{emami2018learning} in 20-cities case which is the only reported TSP size by the authors. Table \ref{tbl:length} also presents the results of NCO \cite{bello2016neural} in Active Search mode, because it is similar to our method. In this mode, the model starts from scratch and searches the space for a predetermined number of steps and are 3.4\% and 3.5\% from the optimal for 20-city and 50-city instances respectively.

The breakdown of the execution time of TauRieL is crucial, because training, sampling, and inference occur in the main loop. Thus, Table \ref{tbl:execution_time} presents the execution times of several steps of the algorithm for 20 and 50-city instances. The figure allows observing the change of execution times of the steps concerning problem size. Increasing from 20-city to 50-city instances do not affect the training and inference time significantly as well as the transition matrix update step.  In addition, the results show that \textit{Sampling} and \textit{Inference} steps have similar execution time. Nevertheless, the sampling step starts to dominate with increasing input.

Similarly, the execution of \textit{Input Gen.} step which is responsible for resizing and appending samples and batches of tensors before starting the search algorithm, increases with increasing data size. The \textit{Sampling} and \textit{CL Merge} costs start to increase when the sample size increases, because at each step of the main loop, the sampling step executes multiple times, and merge step has more sub tours to merge for the global tour. \textit{Others} represents initialization of vectors, the comparison and the update of current best policy and tour length.

In Table \ref{tbl:exec} we compare the training durations NCO \cite{bello2016neural} with TauRieL. TauRieL can generate competitive results per TSP instance in seconds time whereas NCO necessitates hours long training durations before starting inference for a TSP instance. Although NCO presents an online version of their algorithm, the best results are obtained in the offline version that we list in Table \ref{tbl:execution_time}. Specifically, we measure the training times for 20-city and 50-city instances from the reference implementation as 19860 and 36021 seconds respectively. Once the training is finished, the inference is made from the trained model. On the other hand, without needing any data sets, TauRieL runs in 170 seconds for 20-city and 580 seconds for 50-city instances when sample and episodes are both set to 250. Also, for both cases, we obtain 2.4\% and 6.1\% from the optimal compared to 1.4\% and 3.5\% which are reported by \cite{bello2016neural} for 20-city and 50-city instances respectively. We measured the training and inference times of the methods from a reference implementation of \cite{bello2016neural} with uniformly generated random cities in $[0,1] \in \mathbb{R}^2$.

\begin{table}[t]
\caption{Comparison of execution times and tour length gap from optimal between our implementation and NCO \cite{bello2016neural}}
\label{tbl:exec}
\vskip 0.15in
\begin{center}
\begin{small}
\begin{sc}
\begin{tabular}{llll}
\toprule
& Exec. Time (sec) & \multirow{2}{*}{\% from Opt.} & \tabularnewline
 & (Train, Inference) &  & \tabularnewline
\midrule
\multirow{2}{*}{n = 20} & $19860$, 0.04 & 1.4\% & NCO\tabularnewline
 &  170 & 2.6\% & TauRieL\tabularnewline
\midrule
\midrule 
\multirow{2}{*}{n = 50} & $36021$, 0.04 & 3.5\% & NCO\tabularnewline
 &  580 & 6.1\% & TauRieL\tabularnewline
\bottomrule
\end{tabular}
\end{sc}
\end{small}
\end{center}
\vskip -0.1in
\end{table}

\begin{figure}[t]
\centering     
\subfigure[]{\label{fig:tr20}\includegraphics[width=\linewidth]{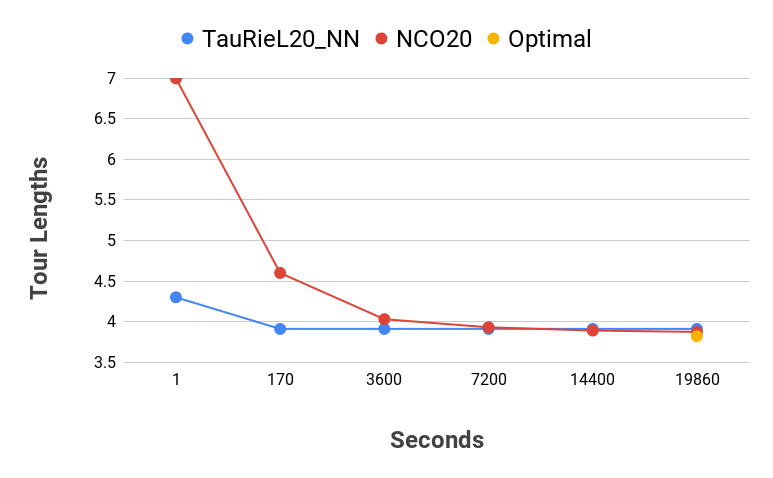}}
\subfigure[]{\label{fig:tr50}\includegraphics[width=\linewidth]{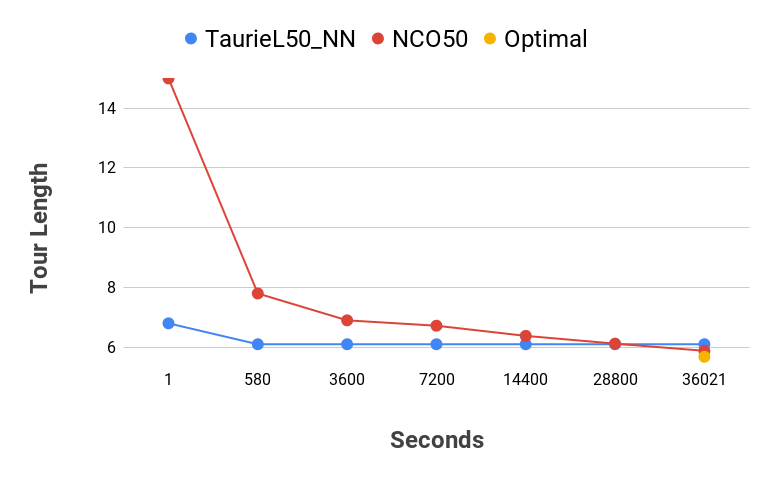}}
\caption{The average tour length vs training duration for 20 and 50-city instances of TauRieL and NCO \cite{bello2016neural}}
\end{figure}

\begin{table}[t]
\caption{The \% gap from 50-city Ptr-Net \cite{vinyals2015pointer} with respect to sample size and the number of training steps}
\label{tbl:change}
\vskip 0.15in
\begin{center}
\begin{small}
\begin{sc}
\begin{tabular}{cccc}
\toprule 
Samples &  & Training Steps & \tabularnewline
 & 50 & 150 & 300 \tabularnewline
\midrule
10 & 18.60 & 17.40 & 15.20\tabularnewline
50 & 15.40 & 10.90 & 8.40\tabularnewline
200 & 10.10 & 8.00 & 7.30\tabularnewline
400 & 8.10 & 7.90 & 2.98\tabularnewline
\bottomrule
\end{tabular}
\end{sc}
\end{small}
\end{center}
\vskip -0.1in
\end{table}

\begin{figure}[t]
\centering     
\subfigure[]{\label{fig:better20}\includegraphics[width=0.99\linewidth]{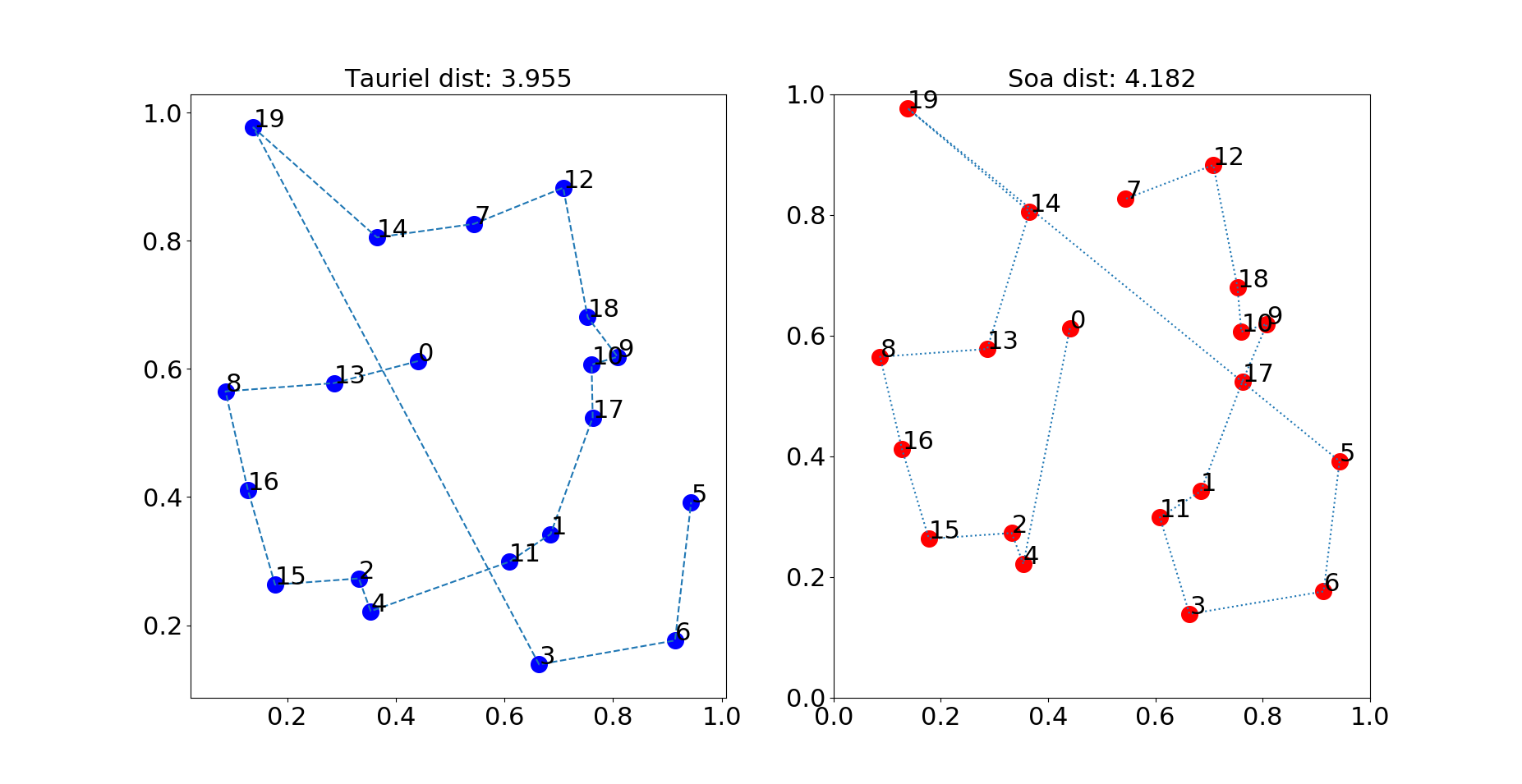}}
\subfigure[]{\label{fig:worse20}\includegraphics[width=0.99\linewidth]{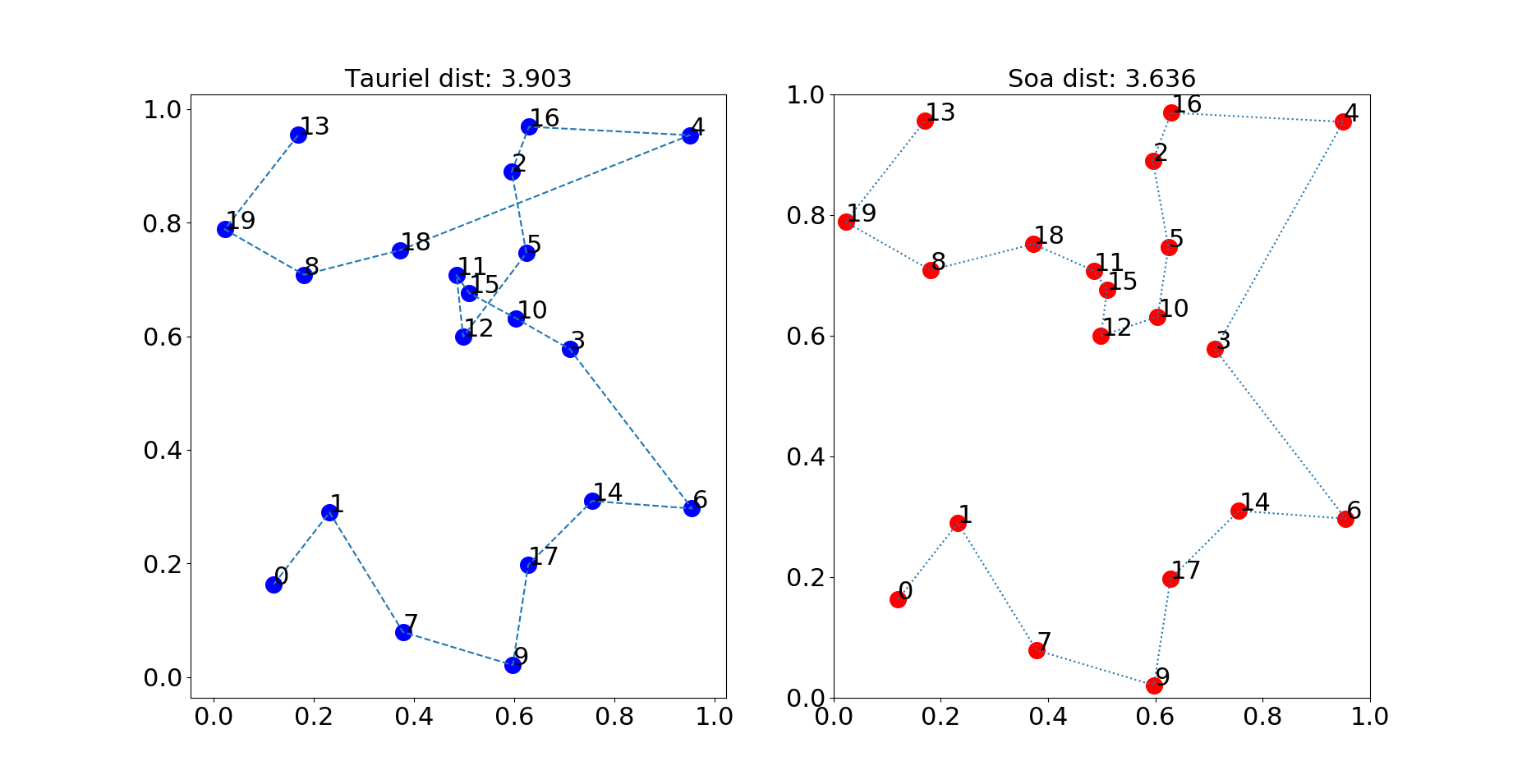}}
\caption{TauRieL's 20-city tour-length results better (a) and worse (b) than the validation set}
\end{figure}

\begin{figure}[t]
\centering     
\subfigure[]{\label{fig:better50}\includegraphics[width=0.99\linewidth]{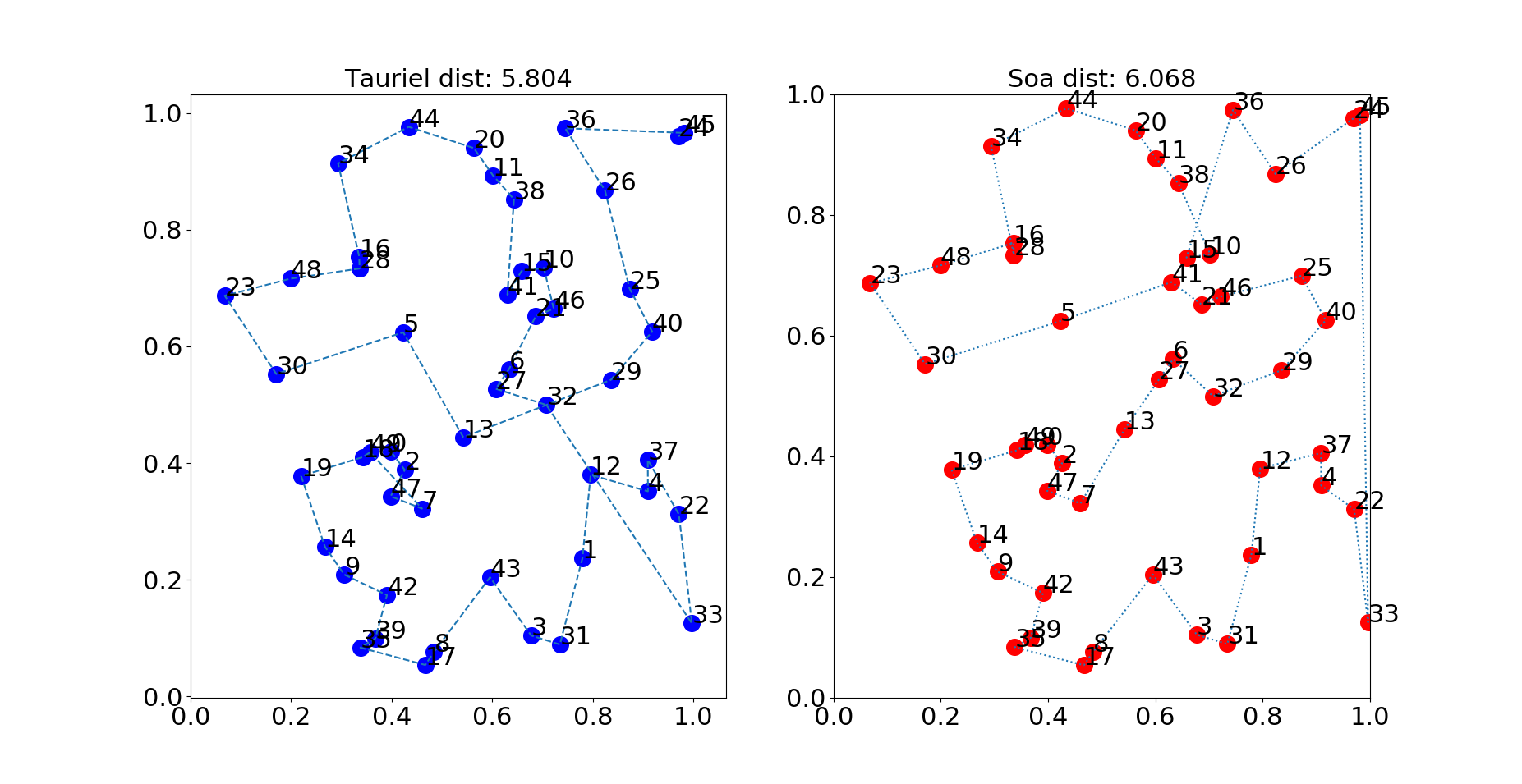}}
\subfigure[]{\label{fig:worse50}\includegraphics[width=0.99\linewidth]{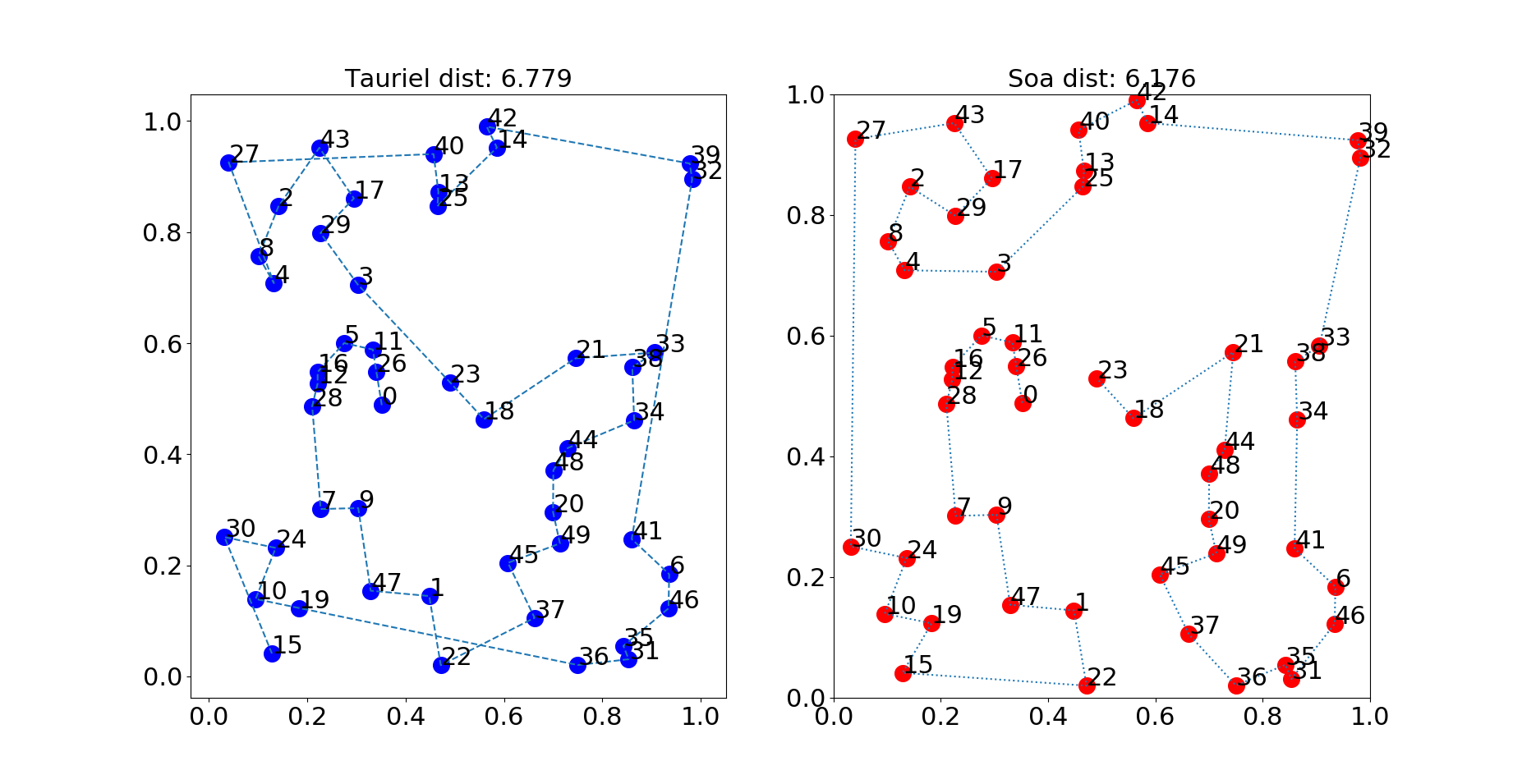}}
\caption{TauRieL's 50-city tour-length results better (a) and worse (b) than the validation set}
\end{figure}

In Figures \ref{fig:tr20} and \ref{fig:tr50}, we introduce the improvements in tour lengths with respect to the training. For 20-city TSP, NCO necessitates more than two hours of training in order to outperform TauRieL which can solve a 20-city instance in less than three minutes. Similarly, for 50-city, NCO needs to train at least eight hours to reach TauRieL's performance whereas TauRieL can obtain a solution in less than ten minutes. The NCO necessitates four hours  and eight hours of training in order to deliver similar solutions for 20-city and 50-city cases respectively.

The sample size and the number of episodes are the most critical parameters for Algorithm \ref{alg:search}. Thus, Table \ref{tbl:change} displays the change of the tour length gap for the sample size and the number of episodes concerning Ptr-Net \cite{vinyals2015pointer}. Keeping a low sample size implies fewer explorations in the design space. After 50 episodes, we observe the benefits of increasing the sampling size. Nevertheless, sampling from the transition also has computational costs. Besides, the number of episodes above 150 have always provided the best results. Thus, in order to reach the best results, our strategy has been to increase the sample size and the number of episodes together.

In Figures \ref{fig:better20} to \ref{fig:worse50}, we compare the tour lengths obtained from TauRieL and validation datasets that are used by Ptr-Net and NCO \cite{vinyals2015pointer,bello2016neural}. Figures \ref{fig:better20} and \ref{fig:better50} are the examples that TauRieL outperforms the given route, and in Figures \ref{fig:worse20} and \ref{fig:worse50} are the examples that TauRieL underperforms. For both methods, there happen long jumps from nearby dense regions to distant points. Because after touring nearby dense regions the algorithm has had to stochastically continue to another candidate city while still maintaining the nearby dense routes. On the other hand, there also exist counterexamples. In Figure \ref{fig:better20}, TauRieL has a longer jump between two cities; however overall tour length is shorter because of better routing at the dense regions.

\section{Related Works}
\label{sec:related_works}
Combinatorial optimization is a branch of mathematical optimization and it has been a cornerstone research field with various application domains from biotech, finance to manufacturing. There are many problems arising in this field that do not yield optimal solutions with polynomial-time algorithms and this constitutes the main reason for further research. A set of reducible problems described by Karp et al. laid the foundations \cite{karp1972reducibility} of combinatorial problems while subsequent research tackled computationally intractable algorithms through approximation methods with performance guarantees \cite{garey2002computers}. 

Optimization problems have caught the attention of machine learning community with recent advances in neural net architectures. Vinyals et al. have proposed pointer nets which are attention based sequence-to-sequence learning architectures and presented results on TSP \cite{vinyals2015pointer}. Their proposed architecture have managed to generate competitive results for other problems such as delaunay triangulation without significant hyperparameter explorations. However, the problem sizes haven't been as large as state-of-the-art \cite{applegate2009certification} which have employed TSP specific local search moves \cite{stutzle2017iterated} and provided optimal solutions up to thousands of elements.

A wide range of TSP variants have been applied to a wide range of domains. Examples include the use of TSP in music for conjunct melody generation in computer-aided composition, as well as for forming automatic playlists and track/artist suggestions - now used by Spotify and Tidal \cite{pohle2005generating}. TSP heuristics have also been used for diffractometer guidance in X-ray crystallography \cite{bland1987large}; and for Telescope scheduling in exoplanet discovery \cite{kolemen2008optimal} as well as in galaxy imaging \cite{carlson1997hazy}. It can also be applied to bioinformatics as a genome ordering problem by posing the ordering as a path traveling through each gene marker \cite{agarwala2000fast} or as a clustering problem to solve gene expression clustering \cite{climer2006rearrangement}.  

Previous research has employed neural nets in order to complement local search heuristics \cite{skubalska2017exploring,wang2017cellular}. One recent example is that authors in \cite{skubalska2017exploring} use Kohonen nets and devise a solver. Self organizing nature of Kohononen net iteratively executes and tries to map neurons which are dispersed onto the 2-D euclidean plane to the cities. At each iteration neurons attempt to decrease the distance between the cities in the neighborhood region which is a parametric set that consists of the nearby cities. The learning rate and the neighborhood function are the hyperparameters. Due to the fact that the number of neurons are higher than the cities, the algorithm presents an ordering mechanism for selecting the best fit neurons that represent the cities.   

Bello et al. \cite{bello2016neural} proposed a TSP solver RL framework based on neural nets. Pointer-nets have been employed for policy gradient and expected tour length prediction. The sequential nature of the framework resembles tour construction algorithms. Stochastic sampling and the actor-critic architecture updates the expected tour length with the current policy (on-policy) and the gradient updates are worked out using reinforce algorithm \cite{williams1992simple}. In addition, a separate neural net for tour exploration incorporate an expected reward based value iteration approximation. All neural nets are constructed using pointer nets \cite{vinyals2015pointer}. Both pre-training and random initialization of the weights are realized as the initial starting state. The framework presents improved tour lengths and execution times compared to Christofides and supervised learning methods \cite{christofides1976worst,vinyals2015pointer}.

The authors of \cite{khalil2017learning} present a deep Q-learning \cite{mnih2013playing} and a graph embedding based solution to target combinatorial optimization problems. Given a problem as graph, they first perform function mappings such as belief propagation that learns feature vectors from latent variable models. Then, the learned embeddings allow to learn a construction graph building heuristic. The Q-learning allows to construct the solution based on the reward which is defined as the change in the cost function among the candidates. A helper function is also employed that helps to satisfy the constraints of the combinatorial optimization from the partial solution throughout the construction process.

Deudon el at. \cite{deudon2018learning} proposes a TSP solver framework based on the attention architecture \cite{vaswani2017attention}. The attention mechanism consists of the input encoder and output decoder. The input encoder uses multi-head attention mechanism to compute the embeddings of the input nodes. Given an input set, the output decoder calculates a probability distribution over the input nodes using attention. The decoder uses the chain rule to factorize the probability of a tour as the mean of all the node embeddings that are generated at each hidden attention layer. The critic net computes a glimpse vector, a weighted sum of the action vectors. The authors also improve the best tour results by using 2-opt heuristic.

The work presented in \cite{emami2018learning} uses an actor-critic based learning on permutation matrices. Authors target combinatorial problems that coincide with graph theory such as Maximum Weight Matching problem (MWM). Authors rely on GRU structures for representation learning over all possible match pairs in a given graph after non-linear embeddings followed by an outer dot product for pairwise correlation mappings. The actor-network maximizes the average reward by choosing the deterministic action $a$ generated by the network and with the reward obtained from the critic net $Q$. 

Furthermore, the critic network optimizes the mean square error between the rounded matrix produced by the Hungarian algorithm \cite{kuhn1955hungarian} and the output of the Sinkhorn layer. The Sinkhorn layer generates a doubly stochastic permutation matrix from the output of the correlation mappings. Similar to the DDPG \cite{lillicrap2015continuous} algorithm, the critic network takes action and a representation of the learned state generated by actor net. TauRieL does not enforce any double stochasticity and can generate permutations from the transition matrix either stochastically or deterministically.

\section{Conclusions}
\label{sec:conc}
In this paper, we present TauRieL. By employing an actor-critic inspired architecture with DNNs, and a state transition matrix, TauRieL can generate 50-city TSP instances in minutes. Solvers that rely on offline training require lengthy training times with the advantage of fast inference. Although heuristics based solutions are still superior to deep learning based solvers regarding accuracy and execution time for larger problem sizes, TauRieL has shown to decrease the execution time gap by two orders of magnitude with comparable solutions compared to recent deep learning based solvers. 

\balance
\bibliographystyle{icml2019}
\bibliography{icmlbib_2}
\end{document}